\PassOptionsToPackage{prologue,dvipsnames}{xcolor}
\PassOptionsToPackage{unicode}{hyperref}
\PassOptionsToPackage{naturalnames}{hyperref}
\documentclass[sigconf,natbib=true,anonymous=false]{acmart}
\renewcommand\footnotetextcopyrightpermission[1]{} % removes footnote with conference information in first column
\usepackage{xspace} %  
\usepackage{algorithm}
\usepackage{algpseudocode}
\usepackage{svg}
\usepackage{float}
\usepackage[dvipsnames]{xcolor}
\usepackage{underscore}
\usepackage[english]{babel}
\usepackage{amsmath}
\usepackage[most]{tcolorbox} % define prompt box
\newtcolorbox[list inside=prompt,auto counter]{prompt}[1][]{
    colbacktitle=black!60,
    coltitle=white,
    fontupper=\footnotesize,
    boxsep=5pt,
    left=0pt,
    right=0pt,
    top=0pt,
    bottom=0pt,
    boxrule=1pt,
    #1,
}

\def\oreo{\emph{\textbf{\textit{Oreo}}} }
\AtBeginDocument{%
  \providecommand\BibTeX{{%
    Bib\TeX}}}

\setcopyright{none}

\usepackage{ragged2e}
\usepackage{booktabs, makecell, multirow}
\usepackage{amsthm}
\usepackage{siunitx}
\usepackage{colortbl}
\usepackage{pifont}
\usepackage{makecell}
\usepackage{subfloat}
\usepackage{caption}
\usepackage{xcolor}
\usepackage{subcaption}
\AtBeginDocument{%
  \providecommand\BibTeX{{%
    \normalfont B\kern-0.5em{\scshape i\kern-0.25em b}\kern-0.8em\TeX}}}

\begin{document}

%%
%% The "title" command has an optional parameter,
%% allowing the author to define a "short title" to be used in page headers.
%\title{Enhancing RAG by Transforming Retrieved Context}
\title{\includegraphics[height=20pt,width=20pt]{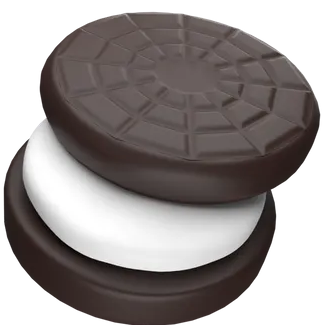} Oreo: A Plug-in C\underline{o}ntext \underline{Re}c\underline{o}nstructor to Enhance Retrieval-Augmented Generation }
%%
%% The "author" command and its associated commands are used to define
%% the authors and their affiliations.
%% Of note is the shared affiliation of the first two authors, and the
%% "authornote" and "authornotemark" commands
%% used to denote shared contribution to the research.

\author{Sha Li}
\affiliation{%
  \institution{Virginia Tech}
  \city{Blacksburg}
  \country{USA}}
\email{shal@vt.edu}

\author{Naren Ramakrishnan}
\affiliation{%
  \institution{Virginia Tech}
   \city{Arlington}
   \country{USA}}
\email{naren@vt.edu}

%%
%% By default, the full list of authors will be used in the page
%% headers. Often, this list is too long, and will overlap
%% other information printed in the page headers. This command allows
%% the author to define a more concise list
%% of authors' names for this purpose.
\renewcommand{\shortauthors}{Li et al.}

%%
%% The abstract is a short summary of the work to be presented in the
%% article.
\begin{abstract}
Retrieval-Augmented Generation (RAG) aims
to augment the capabilities of Large Language Models (LLMs) by retrieving and incorporate external documents or chunks prior to generation. 
However, even improved retriever relevance can brings erroneous or contextually distracting information, undermining
the effectiveness of RAG in downstream tasks. We introduce a compact, efficient, and pluggable module designed to refine retrieved chunks before using them for generation. The module aims to extract and reorganize  the most relevant and supportive information into a concise, query-specific format. Through a three-stage training paradigm---comprising supervised fine-tuning, contrastive multi-task learning, and reinforcement learning-based alignment---it prioritizes critical knowledge and aligns it with the generator’s preferences. This approach enables LLMs to produce outputs that are more accurate, reliable, and contextually appropriate.

\end{abstract}

%%
%% Keywords. The author(s) should pick words that accurately describe
%% the work being presented. Separate the keywords with commas.

\settopmatter{printfolios=true} 
\maketitle
\pagestyle{plain}
\section{Introduction}
\label{intro}
Large language models (LLMs) have demonstrated remarkable versatility across a wide spectrum of natural language processing (NLP) tasks, subsuming pipelines that were originally tailormade for each task. Despite being trained on massive text corpora, LLMs still face memory-related challenges such as out-of-date and out-of-domain knowledge, and they occasionally hallucinate non-factual or non-sensical content~\cite{zhou21aclfindings, maynez-etal-2020-faithfulness}. To enhance the accuracy and reliability of LLM-generated outputs, retrieval-augmented generation (RAG) has emerged as a promising solution for knowledge-intensive tasks \cite{zhu2023large, gao2023retrieval, li2022survey} (\textit{eg.}, open-domain question and answering).
RAG systems typically follow a \textit{``retrieve-then-generate"} paradigm \cite{shao-etal-2023-enhancing}, where a \textit{retriever} identifies relevant information from an external corpus and uses this information to augment
context in constituting the input to a generative model (\textit{ie.}, \textit {generator}), thus yielding an improved answer. 

\begin{figure}[H]
\begin{center}
  \includegraphics[scale=0.44]{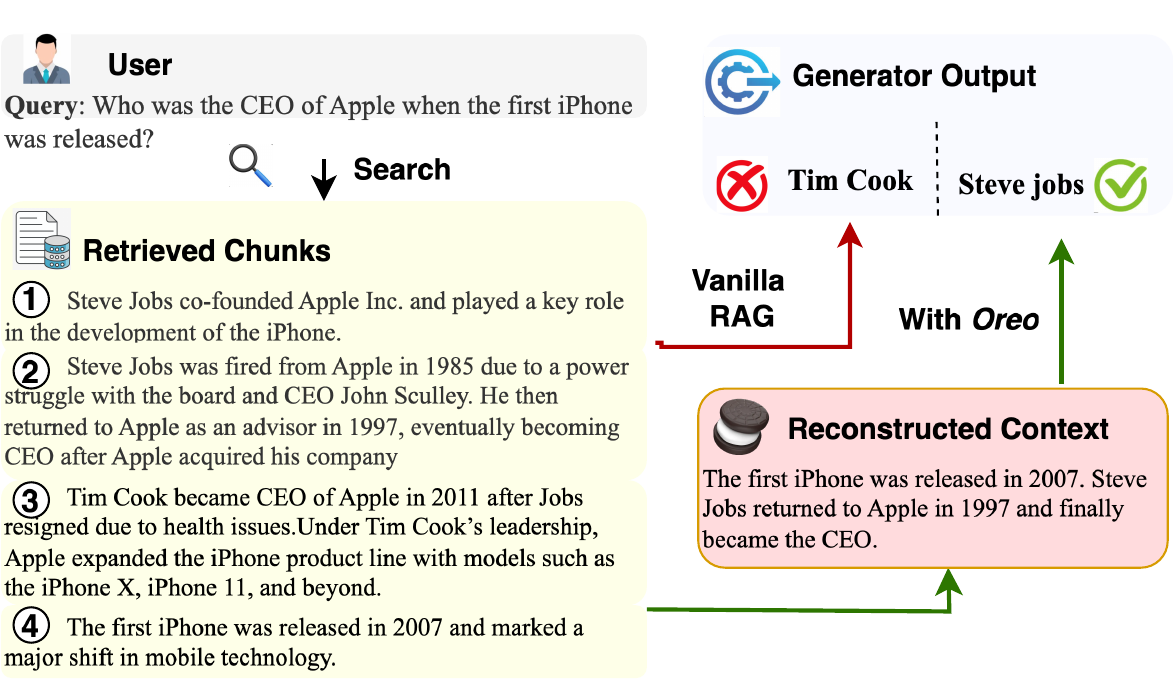}
  \end{center}
  \caption{\textmd{An example comparing {\color{red}{vanilla RAG}} versus {\color{ForestGreen}{RAG with \oreo}} highlights the impact of redundant and scattered information within the retrieved document chunks. In the vanilla RAG setup, even though the retrieved chunks contain contextually relevant information to the query, the presence of distractions and redundancy misleads the downstream LLM, causing it to misinterpret temporal dependencies and generate an incorrect answer. In contrast, \oreo effectively captures the essential evidence and reconstructs the context, leading to accurate and correct responses.}}
  \Description{introimage}
  \label{fig:intro}
\vspace{0.6em}
\end{figure}

Despite its promise, a vanilla RAG system usually comes with shortcomings that can hinder its effectiveness. One major issue is semantic dissonance between the user query, the retriever, and the generator. This occurs when the retrieved documents, while semantically or contextually related to the topic, fail to directly address the query, leading to suboptimal answers~\cite{cuconasu2024power, wu2024easily}. Another challenge pertains to the presence of noise, \textit{ie.} misleading, redundant, distracting, or even erroneous information within the retrieved documents. Such noise can misguide the generator, resulting in inaccurate or incoherent answers~\cite{sun2023contrastive, shi2023large}. For complex tasks that necessitate reasoning across multiple documents, the generator often struggles to correlate dependencies and relationships between them \cite{behnamghader2022can}, leading to reasoning errors. For example, as illustrated in Figure \ref{fig:intro}, the correct answer while present in the retrieved chunks is
not captured by the vanilla RAG process.
Additionally, RAG systems are prone to the ``lost-in-the-middle''\cite{liu2023lost} dilemma, where LLMs exhibit a tendency to prioritize information presented at the beginning and end of an input sequence, while paying less attention to the middle. Finally, the lack of joint optimization between the retriever and the generator exacerbates issues such as \textit{knowledge inconsistencies} \cite{wang-etal-2023-causal} or \textit{knowledge conflicts} \cite{xu-etal-2024-knowledge-conflicts}, which prevent the generator from producing accurate and contextually appropriate responses as the retrieved knowledge fails to adequately support the generation.

To address these challenges, many solutions have been proposed in prior research. Techniques such as query decomposition \cite{chan2024rq, kim-etal-2023-tree}, query rewriting \cite{wang-etal-2023-query2doc,tan2024small, chan2024rq, ma-etal-2023-query}, and query expansion \cite{lei-etal-2024-corpus} aim to improve retriever performance by refining or enriching the input queries. Some studies have integrated rerankers \cite{yoran2023making, yu2023chain, nogueira2019multi} into retrieval systems,  which reorder and prioritize the most relevant documents to ensure that the most pertinent information is provided to the generator. These works attempt to optimize the context on the passage level and largely ensure relevance with the query, but they still face challenges in maintaining comprehensive attention to the nuanced, finer-grained details of query-specific information. 

Further advancements have been made in noise and redundancy exclusion. For example, filters based on lexical and information-theoretic approaches have been developed to identify and preserve useful content while directly eliminating less relevant information \cite{wang2023learning, li-etal-2023-compressing, jiang-etal-2024-longllmlinguaCompAct}. Summarization techniques  \cite{xu2024recomp} have been developed to synthesize and condense query-focused information from retrieved documents, leveraging extraction or abstraction methods. Compression techniques \cite{chevalier-etal-2023-adapting, cao-etal-2024-retaining, cheng2024xrag, yoon-etal-2024-compact, jiang-etal-2023-llmlingua, pan-etal-2024-llmlingua, jiang-etal-2024-longllmlinguaCompAct} extend this functionality by generating summary vectors that encode essential information for downstream tasks. While these methods improve efficiency, they do not align the retriever and generator in a manner that guarantees effective collaboration, which often result in knowledge gaps and consequently incorrect or suboptimal generation. From a training perspective, concurrent \cite{guu2020retrieval, lin2023ra, zamani2024stochastic, izacard2022few} or asynchronous \cite{zhang-etal-2024-arl2, shi-etal-2024-replug} training of retrievers and generators is a widely adopted strategy to improve their interaction and collaboration \cite{guu2020retrieval, borgeaud2022improving}. Although such techniques foster synergistic improvements, they can be computationally expensive and often require large amounts of annotated data to achieve optimal results.

In this work we introduce \oreo, a c\underline{O}ntext \underline{RE}c\underline{O}nstructor designed to enhance the performance of RAG systems on knowledge-intensive tasks by \textit{optimizing the quality of context} and \textit{mitigating knowledge inconsistencies}. \oreo is implemented in a plug-and-play manner, functioning as an intermediary module between the retriever and the generator. It receives document chunks from the retriever and produces refined context tailored for the generator. Instead of merely extracting critical tokens from the chunks, \oreo reorganizes them and generates condensed query-aware summaries. Additionally, \oreo synergizes the reconstructed context with the generator’s behavior of knowledge acquisition, ultimately leading to more accurate and contextually relevant answers.

Our key contributions are:
\begin{enumerate}
    \item We propose enhancing the RAG by introducing a ``retrieve-reconstruct-then-generate" paradigm, offering a novel perspective on refining retrieved content for improved integration of external knowledge in RAG. \oreo
    overcomes the lack of contextual integration among fragmented chunks in vanilla RAG by extracting subtle relations from scattered facts, and transforming redundant context into a concise context.
\item \oreo is a plug-and-play module, inherently modular, generalizable, flexible and robust, powered by a three-stage training scheme comprising supervised fine-tuning, contrastive multi-task learning and reinforcement learning. This enables seamless integration with arbitrary retrievers, generators, and off-the-shelf RAG systems. 
\item We demonstrate \oreo's efficiency, effectiveness and robustness for both single-hop QA tasks (PopQA \cite{mallen-etal-2023-trust}, NaturalQuestion (NQ) \cite{kwiatkowski2019natural}, TriviaQA (TQA) \cite{JoshiTriviaQA2017}), and multi-hop QA tasks (HopotQA \cite{yang-etal-2018-hotpotqa}, 2WikiMultiHopQA \cite{ho-etal-2020-constructing}). On average, \oreo contributes 5.115\% downstream performance while reducing the input token length for generator by 12.87x.
\end{enumerate}
%Revisiting the motivational example in Figure \ref{fig:intro}, the lack of  systems often results in disjointed responses or hallucinations. In contrast, \oreo overcomes these limitations by extracting query-specific evidence, capturing , finally  guiding the generator to produce accurate answers. By design, \oreo offers several key advantages. First, \textit{producing condensed and refined context.} \oreo significantly shortens the length of retrieved context by extracting, abstracting, and reorganizing supporting evidence into concise and query-aware context. This enables the generator to focus on the most critical information, improving response quality. Second, \textit{plug-and-play flexibility}. The modular nature of \oreo makes it highly adaptable, allowing seamless integration with arbitrary retrievers, generators, and off-the-shelf RAG systems. This flexibility enables broad applicability across diverse RAG configurations without significant adjustments. Third, \textit{lightweight and efficient}. \oreo is lightweight and trainable, enhancing system performance while maintaining computational feasibility.

%\textbf{Contributions. } (i) (ii)  (iii) We demonstrate improved performance and reduced token length for factual short-form QA tasks. 

\section{Methodology}
\label{method}
A typical RAG system comprises of two primary components that work in tandem: the retriever $\mathcal{R}$ identifies and retrieves top-\textit{k} document chunks $\mathcal{D}=\{d_{1}, d_{2},...,d_{k}\}$ from an external knowledge base based on their relevance to a given query \textit{q}; the generator $\mathcal{G}$ then produces the final answer for \textit{q} by conditioning on the combination of $\mathcal{D}$ and query \textit{q}, formally expressed as $y=\mathcal{G}(\mathcal{D}, q)$. However, the performance of such a general pipeline is compromised by reasons we indicated in \S\ref{intro}. Therefore, we propose \oreo to reconstruct context by extracting the most supportive evidence from $\mathcal{D}$, and producing a concise, query-aware context $\mathcal{C}$ that aligns with the knowledge acquisition mechanics and preference of $\mathcal{G}$. An ideal $\mathcal{C}$ should be produced after \oreo identifies essential entities and facts from $\mathcal{D}$, establishes their relations, and retains only the necessary information for $\mathcal{G}$ to effectively answer \textit{q}. This process goes beyond plain information extraction, as it involves organizing and synthesizing content into a coherent and query-specific context. Therefore, we formulate the context reconstruction task as (itself) a text generation problem.

\subsection{Method Overview}
\label{met_overview}
\begin{figure}[htbp!]
\begin{center}
  \includegraphics[scale=0.54]{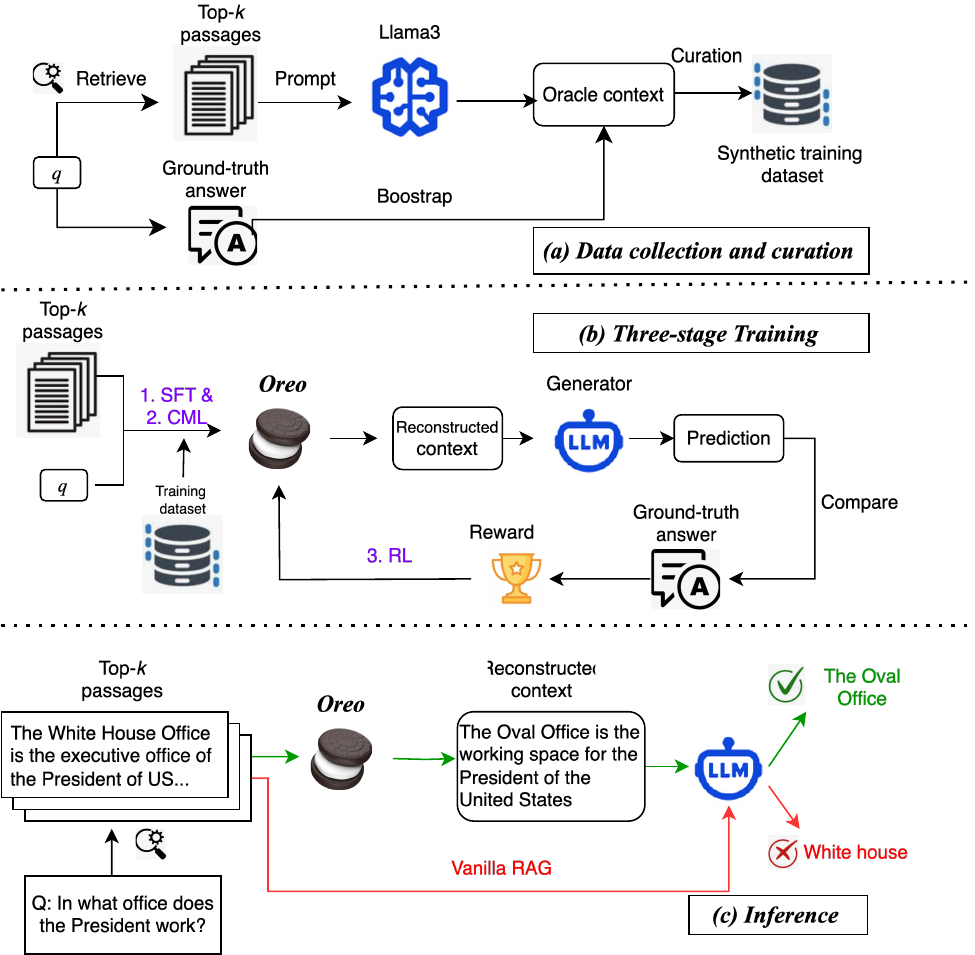}
  \end{center}
  \caption{\textmd{The framework of \oreo. (a) outlines the process of data collection and curation (\textbf{top}). (b) demonstrates the three-stage training, which comprises the supervised fine-tuning (SFT), contrastive multi-task learning (CML) and reinforcement learning (RL) alignment (\textbf{middle}). (c) illustrates the application of \oreo, comparing against the vanilla RAG (\textbf{bottom}).}}
  \Description{framework}
  \label{fig:overview}
    \vspace{0.5em}
\end{figure}

Our method extends the standard RAG paradigm from ``retrieve-then-generate'' to ``retrieve-reconstruct-then-generate''. Specifically, we train a text generation model $\mathcal{M}_{\theta}$, parameterized by $\theta$ to map the retrieved documents $\mathcal{D}$ into a refined context \textit{c} that enables the downstream generator $\mathcal{G}$ to produce more accurate answers for an input query: $c=f_{\mathcal{M}_{\theta}}(\mathcal{D}, q)$. The training of $\mathcal{M}_{\theta}$ involves three stages: 1. $\mathcal{M}_{\theta}$ is trained to learn the transformation from original documents to refined context using annotated datasets (\S\ref{sft}). 2. Self-generated samples are incorporated to enhance the model's ability to recognize and correct its own errors, thereby improving robustness and generalization (\S\ref{cl}). 3. The reconstructed context is aligned with the generator’s knowledge acquisition process by incorporating feedback from $\mathcal{G} 
  $(\S\ref{rl}). However, obtaining an annotated dataset with refined context for SFT is challenging. Drawing inspiration from prior work \cite{balachandran-etal-2022-correcting}, we replace human annotation with advanced LLMs to generate high-quality synthetic oracle training data (\S\ref{data_coll}). An overview of the framework is depicted in Figure \ref{fig:overview}.

\subsection{Data Collection and Curation}
\label{data_coll}
\textbf{Data collection}. To train \oreo during the SFT stage, an annotated dataset containing context with the most supportive evidence from retrieved document chunks is crucial. Such context should be query-specific, answer-aware, grounded in retrieved chunks, and structured as a rationale chain capable of deriving the correct answer. However, such datasets are not readily available, and manually annotating evidence for each query is time-consuming and labor-intensive. Fortunately, contemporary LLMs have exhibited impressive instruction learning capabilities to extract useful information \cite{dagdelen2024structured} and generate high-quality reasoning steps \cite{wei2022chain} even in few-shot settings \cite{brown2020language}. In this work, we elicit such capability from more advanced LLMs to our relatively smaller model \oreo, through generating a high-quality reasoning dataset using LLMs and utilizing it as ``gold context" to train \oreo. Specifically, given a query and corresponding retrieved document chunks, we first prompt Llama3-8B-Instruct \cite{touvron2023llama} to extract key entities and events from $\mathcal{D}$, and generate detailed rationales to answer the query. Since we prioritize the information extraction capability of \oreo during the SFT stage, to ensure reliability and minimize hallucinations, we construct the gold training dataset solely from query-document pairs where the ground-truth answer is explicitly present within the retrieved chunks.

\textbf{Bootstrapping}. For queries where the generated reasoning fails to include the ground truth (despite it being present in the retrieved chunks), we bootstrap Llama3 by providing the correct answer and iteratively reprompting it to perform generation. Such an iterative process allows Llama3 to reason backwards and learn to generate rationale chains that support the correct answer. This bootstrapping process is inspired by \cite{zelikman2022star, wei2024instructrag}. The prompts and demonstrations used for gold context generation and boostrapping are provided in Appendix~\ref{app:prompts}.

\textbf{Data curation}. Accurate extraction of supporting evidence and reasoning from query to answer is essential for training $\mathcal{M_{\theta}}$. To eliminate hallucination and ensure the quality of learning, we conduct data curation by applying the following rules. \textit{1. Ground Truth Alignment.} We retain query-context pairs where the generated context from Llama3 contains ground truth answers. \textit{2. Entity and Event Consistency.} We extract sets of entities and events from both the original documents and the Llama3-generated context. Instances are retained only if the entities and events extracted from the generated context ($\mathcal{E}_{gen}$) are a subset of those present in the original documents ($\mathcal{E}_{ori}$). By following these steps, the refined context generated by Llama-3 is treated as ``gold context" for training $\mathcal{M_{\theta}}$.

\subsection{Supervised Fine-tuning}
\label{sft}
With the curated dataset constructed in \S\ref{data_coll}, we employ supervised fine-tuning (SFT) to elicit the ability of extracting and reasoning from LLM to \oreo. Specifically, given a curated training dataset \ $\mathcal{T}=\{{x}_{i}, c_{i}\}^{N}_{i=1}$, where $x_{i}$ is the combination of query $q_{i}$, the associated retrieved document chunks $\mathcal{D}_{i}$ and task instructions. The goal of SFT is to train a sequential model $\mathcal{M}_{\theta}$ to generate target context conditioned on the $x$, and preceding tokens $c_{<t}$ of the context. The model minimizes the negative log-likelihood over the gold context, as defined by the following loss function:
\begin{equation}
\begin{split}
\label{eq:sftloss}
\mathcal{L}_{SFT}= &\mathbb{E}_{(x, c)\sim \mathcal{T}}[-log \ p_{\mathcal{M}_{\theta}}(c|x)] \\
 = & - \sum_{t=1}^{L}log\ p_{\mathcal{M}_{\theta}}(c_{t}|x, c_{< t})
\end{split}
\end{equation}
where \textit{p} represents probability distribution of generation by the model $\mathcal{M}_{\theta}$. 

\subsection{Contrastive Multitask Learning}
\label{cl}
The SFT in \S\ref{sft} serves as the initial step in equipping \oreo with the capability to reconstruct context. By emulating the behavior of an LLM, SFT enables \oreo to extract critical entities, events, and facts from $\mathcal{D}$, capture subtle relationships and organize them into coherent reasoning paths. This process ensures that the reconstructed context effectively supports the generation of accurate and complete answers for queries. However, autoregressive models trained solely on ground truth data often demonstrate suboptimal generalization performance. To address this issue, our goal is broader: we seek to empower \oreo to identify its own errors and integrate sequence-level supervised signals, which are critical for enhancing conditional text generation into training, thus improving its generalization. To achieve this goal, we introduce contrastive learning as a complementary step following SFT.

\textbf{Construct contrastive samples. }Inspired by \cite{an2022cont}, in addition to using in-batch instances, we gather contrastive samples from \oreo's own predictions. Specifically, we obtain the model’s top-\textit{n} recent predictions via beam search, rank and label them as positive and negative pairs ($c^{+}, c^{-}$) in descending order of sequence-level similarity with the gold context $\mathcal{C}$, using the ROUGE metric to measure the similarity.

\textbf{Margin-based pairwise loss.} To guide the learning process, we employ a pairwise margin-based loss that encourages \oreo to bring positive candidates closer semantically to the retrieved document chunks $\mathcal{D}$ while distancing negative ones. This ensures that the positive candidates generated by \oreo capture the essential and grounded information from $\mathcal{D}$ with the guidance of gold context, while discarding irrelevant information. The pairwise loss function is combined with the negative log-likelihood loss from SFT, forming a multi-task learning process. The final loss function is expressed as:
\begin{equation}
\begin{split}
\label{clloss}
    \mathcal{L}_{CL}=& \sum \ max \{0, cos(E_\mathcal{D},E_c^-)  -cos(E_\mathcal{D},E_c^+) \\
    & + \eta *(rank_{c^-}-rank_{c^+})\}  \\
    & + \alpha \mathcal{L}_{SFT}
\end{split}
\end{equation}
where $E$ denotes the vector representations, $\eta$ is the hyperparameter and $rank_{c^+/c^-}$ denotes the ranking position of the candidates respectively, meaning that the contrastive pair with a larger ranking gap should have a larger margin \cite{an2022cont, zhong-etal-2020-extractive}. 

\subsection{Reinforcement Learning Alignment}
\label{rl}
The supervised fine-tuning and contrastive multitask learning stages equip \oreo with the ability to capture critical evidence and retain supportive information from retrieved content. However, knowledge inconsistencies among the retriever $\mathcal{R}$, \oreo and the generator $\mathcal{G}$ persist due to their independent optimization processes. Additionally, training \oreo with keeping $\mathcal{G}$ as a black-box precludes gradient back-propagation from $\mathcal{G}$ to update \oreo. To address these challenges, we incorporate reinforcement learning (RL) into \oreo's training pipeline following the above training stages. This step enables \oreo learn from labeled ground truth of downstream tasks by aligning their output with the needs of $\mathcal{G}$ to produce correct answers. Specifically, we model $\mathcal{G}$ as a reward model and leverage the discrepancy between $\mathcal{G}$'s generated output and ground truth as reward signals. Proximal Policy Optimization (PPO) \cite{ouyang2022training, Stiennon2020LearningTS} is employed to optimize \oreo in this alignment stage. 

\textbf{Policy formulation and optimization}. In this step, $\mathcal{M}_{\theta}$ serves as the policy $\pi_{\theta}$. It takes the reconstructed context $\hat{c}$ from prior training steps and returns a new $\hat{c}^{'}$, optimized by feedback from $\mathcal{G}$. The action space consists of all tokens in the corpus. At each step, the parameterized policy $\pi_{\theta_{t}}$ selects an action $a_{t}$ in a given state $s_{t}$ to maximize the discounted long-term reward $\mathbb{E}_{\pi_{\theta_{t}}}[\sum_{t-0}^{T}\gamma^{t}\mathcal{R}(s_{t}, a_{t})]$. Specifically, the action $a_{t}$ is predicting the next token, and state $s_{t}$ is the sequence of all preceding tokens. The objective function is:
\begin{equation}
\begin{split}
\label{ppoclip}
\mathcal{L}_{RL} = &\mathbb{E}[min(r_{t}(\theta)\cdot A_{t}, clip(r_{t}(\theta), 1-\epsilon, 1+\epsilon)\cdot A_{t})] \\
& - \beta(V(s_{t})-R_{t})^2
\end{split}
\end{equation}
where $r_{\theta}=\frac{\pi_{\theta}(a_{t}|s_{t})}{\pi_{\theta_{old}}(a_{t}|s_{t})}$ is the ratio of the updated policy $\pi_{\theta}$ to previous policy $\pi_{\theta_{old}}$. PPO ensures stable and efficient updates by clipping policy ratios, preventing excessively large changes that could destabilize training. The parameter $\epsilon$ defines how much the new policy can deviate from the old policy. $A_{t}$ is the advantage function, measures
whether or not the action is better or worse than the policy’s old behavior, estimated using Generalized Advantage Estimation (GAE) \cite{schulman2015high}: $A_{t}=\sum_{l=0}^{L}(\gamma\lambda)^{l}(R_{t+l}+\gamma V(s_{t+l-1})-V(s_{t+l}))$ where $\gamma$ and $\lambda$ are discount factors. $V(s_{t})$ is a critic network estimating the value of state $s_{t}$. $R_{t}$ is the estimated reward at time $t$. $\beta(V(s_{t})-R_{t})^2$ weighted by $\beta$ minimizes the discrepency between estimated and true values. 

\textbf{Reward estimation}. With the downstream generator $\mathcal{G}$ serving as a reward model, the generation of \oreo by policy $\pi_{\theta_{t}}$ is passed to $\mathcal{G}$ with query $q$ to generate the answer $y$. When the end of sentence (\textit{eg.}, <EOS>) token is generated, the corresponding reward $R_{t}$ is obtained by comparing the generated answer $y$ with ground truth answer $y_{gold}$, which is measured by the ROUGE score $R_{t}=ROUGE(y, y_{gold})$. However, $\mathcal{G}$ generates answers only after completing all tokens, but (\ref{ppoclip}) updates every action step.  To address this, we incorporate a token-level weighting mechanism \cite{yang-etal-2023-prca}. Considering that a token $t$ with higher generation probability deemed more critical by the current policy. Consequently, the token's contribution to the final reward is proportionally adjusted. We estimate the reward at each step $t$ using the formulation: 
\begin{equation}
\label{rewd}
    R_{t}=\textrm{ROUGE}(y, y_{gold})*\log(\textrm{softmax}(e^{\pi_{\theta})(a_{t}|s_{t})})
\end{equation}
Since the rewards estimated by (\ref{rewd}) are sequence-level and sparse, following \cite{wu2021recursively}, we regularize the reward function using a token-level KL penalty to prevent the model from deviating too far from the initialized LM. The final regularized reward estimation is:
\begin{equation}
\label{regrewd}
    \hat{R_{t}}=R_{t}-\delta KL(\pi_{\theta_{t}}(a_{t}|s_{t})||\pi_{0}(a_{t}|s_{t}))
\end{equation}

\section{Experiments}
We evaluate \oreo across five open-domain question-answering (ODQA) tasks, comparing its performance against a suite of baselines. Our experiments holistically assess the quality of reconstructed context by \oreo along five critical dimensions: \textbf{efficiency}, \textbf{effectiveness}, \textbf{robustness}, \textbf{faithfulness} and \textbf{completeness}. %To this end, we employ a diverse set of task-specific and model-agnostic metrics to assess both system-level performance and the reliability of downstream QA outputs. Additionally, we leverage LLM as evaluators\cite{zhang2023evaluating} to facilitate fine-grained judgment of contextual coverage and factual consistency. 
Our primary emphasis is on \textbf{short-term factual QA tasks}, where the answers are typically concise in a few tokens. These tasks are sensitive to context quality and require precise evidence identification and summarization, making them an ideal benchmark for evaluating the performance of \oreo. In this section, we provide details of tasks and datasets (\S\ref{sec:dst}), baselines (\S\ref{sec:basline}) and experiment setup (\S\ref{sec:expset}). 

\subsection{Datasets and Tasks}
\label{sec:dst}
\textbf{Datasets. } We evaluate \oreo on both single-hop and multi-hop open-domain question answering tasks. For single-hop QA, we use PopQA (PQA) \cite{mallen-etal-2023-trust}, NaturalQuestions (NQ) \cite{kwiatkowski2019natural}, and TriviaQA (TQA) \cite{JoshiTriviaQA2017}. For multi-hop QA, we test \oreo on the more complex HopotQA (HQA) \cite{yang-etal-2018-hotpotqa} and 2WikiMultiHopQA (2WQA) \cite{ho-etal-2020-constructing}, where each question requires reasoning over multiple articles.

%We employ PopQA (PQA) \cite{mallen-etal-2023-trust}, NaturalQuestions (NQ) \cite{kwiatkowski2019natural}, and TriviaQA (TQA) \cite{JoshiTriviaQA2017} to experiment with the single-hop open-domain QA task. Each sample in these datasets has a question and annotated short extractive answers. For PopQA, we utilize the long-tail subset, which comprises 1,399 queries involving rare entities with fewer than 100 monthly Wikipedia page views \cite{asai2023self}. \textbf{Multi-hop question answering. }We also test \oreo on more complex QA scenarios, specifically using multi-hop datasets HopotQA (HQA) \cite{yang-etal-2018-hotpotqa} and 2WikiMultiHopQA (2WQA) \cite{ho-etal-2020-constructing}. Each question in these datasets requires reasoning over multiple articles. 

\textbf{External knowledge source. }For all experiments, we use the Wikipedia dump \cite{karpukhin-etal-2020-dense} as the external knowledge source.  

\textbf{Evaluation metrics. } Following previous studies, \textit{eg.},~\cite{wang2023learning}, we assess extractive QA performance (PopQA, NQ, and TriviaQA) using the Exact Match (EM) metric, while abstractive QA performance (HotpotQA and 2WQA) are measured using unigram F1.

We provide detailed statistics and experimental setups for each dataset in Appendix~\ref{app:stats_dataset}.

\subsection{Baselines}
\label{sec:basline}
For comparison, we focus on evaluating how effectively \oreo enhances vanilla RAG systems treating both the retriever and generator as black-box components, acknowledging that they may be imperfect and not allowed to be fine-tuned. We compare the performance of downstream tasks using five configurations: 
\begin{enumerate}
    \item \textbf{Query only}. The answer generation is performed by using only the query without incorporating any retrieved context. This mostly relies on the internal knowledge of LMs
    \item \textbf{Original full content}. The context for answer generation is the sequential concatenation of all retrieved document chunks. This setup uses raw, unprocessed retrieval results
    \item \textbf{Passage-level filtering.} Only the most relevant chunks are selected as context. Specifically, the chunk that is best-ranked is chosen for each query. For single-hop tasks, only one passage is selected, while for multi-hop tasks, two passages are used
    \item \textbf{Extraction and compression}. We employ eight state-of-the-art information extraction and compression methods to select informative sentences and generate concise summaries from retrieved documents. Specifically:
    \begin{enumerate}
        \item CXMI-trained model, following \cite{wang2023learning}, uses conditional cross-mutual information (CXMI) \cite{fernandes-etal-2021-measuring} to train a language model to filter redundant context by quantifying each sentence’s contribution to the correct answer
    \item Selective-Context \cite{li2023unlocking} removes uninformative content based on self-information
    \item LLMLingua \cite{jiang-etal-2023-llmlingua} and LLMLingua-2~\cite{pan-etal-2024-llmlingua} apply perplexity-based compression to retain sentences that most enhance answer likelihood
    \item  xRAG \cite{cheng2024xrag} encodes passages into a single embedding token, integrating them via modality fusion into the LM’s representation space
\item  CompAct \cite{yoon-etal-2024-compact} is a progressive compression framework that preserves query-aware content
\item EXIT \cite{hwang2024exit} employs an adaptive extractive pipeline to select context based on query relevance
\item  Refiner \cite{li-etal-2024-refiner} uses a decoder-only LLM to extract verbatim query-relevant spans, organizing them by interdependencies
    \end{enumerate}
    \item \textbf{reconstructed context by using \oreo}
\end{enumerate}

\subsection{Experiment setup}
\label{sec:expset}
\textbf{Retriever}.
To retrieve top-\textit{k} document chunks for each query ($k=5$ unless otherwise specified), we employ a range of off-the-shelf retrievers, including  Contriver \cite{lei-etal-2023-unsupervised}, DPR \cite{karpukhin-etal-2020-dense} and BM25 \cite{robertson1994some}. The choice of multiple retrievers ensures that the robustness of \oreo is tested against various retrieval mechanisms, each with different strengths and weaknesses. Additionally, we extend our experiments to include retrieval of the top-\textit{10} document chunks for 2WikiMultihopQA to examine whether \oreo's performance is sensitive to context length. 

\textbf{Downstream generator}. 
We access how do the contexts generated by different methods described in \S\ref{sec:basline} affect the downstream generator by evaluating the performance of QA tasks. Specifically, we use FLAN-T5 \cite{chung2024scaling} and OPT-IML \cite{iyer2022opt} as the downstream generator. (Note that, \oreo operates as an independent module, making it compatible with various retrievers, generators, and other existing RAG frameworks. )

\textbf{Model and training}. We employ T5-small \cite{raffel2020exploring} as the backbone model for \oreo, though it is applicable to any encoder-decoder and auto-regressive models such as LLaMA. \oreo is implemented based on Transformer library \cite{wolf-etal-2020-transformers}, with RL implementation built upon the open-sourced package RL4LM \cite{ramamurthy2022reinforcement}. For CML, we allow a maximum of 12 contrastive samples generated by \oreo and set beam size as 8. Unless otherwise specified, \oreo is trained for 5 epochs during SFT and 3 epochs for CML stages, with batch size 4/8/16 based on the dataset size, and using a learning rate of $5e-5$. Detailed parameter settings are listed in Appendix~\ref{app:params}.

\textbf{Inference}. During inference, we perform ablation studies by varying the number of tokens produced by \oreo to assess its impact on performance.

\begin{table}[H]
\vspace{-1em}
\centering
\caption{\textmd{Parameter settings for experiments. Parameters without being specified are set to their default values as defined by the development package. }}
\begin{tabular}{c|c} 
 \hline
 Parameter & Value \\  
 \hline
    $\eta$ (CML) & 0.01\\
    $\alpha$ (CML) & 0.5\\
    $\epsilon$ (RL) & 0.2 \\
    $\gamma$, $\lambda$(RL) & 0.95 \\
    Top-\textit{k} (RL) & 4 \\
    Top-\textit{p} (RL)& 0.95 \\
 \hline
\end{tabular}
\label{table:params}
\end{table}
\section{Results and Analysis}
We seek to answer the following questions through experiments:
\begin{enumerate}
\item How does \oreo perform in the RAG pipeline for QA tasks compared to alternative context configurations outlined in §\ref{sec:basline}?  (\S\ref{sec:overall})
\item To what extent does \oreo reduce input token length and balance inference latency while improving QA performance? (\S\ref{sec:latency})
\item How robust is \oreo to noisy contexts and perturbations in chunk order? (§\ref{sec:robust})
\item How well does \oreo generalize to out-of-distribution datasets unseen during training? (\S\ref{sec:zeroshot})
\item How effective is \oreo in generating context that are complete to the query and faithful to the retrieved chunks? (\S\ref{sec:qwen_eval})
\item How does the number of tokens in the reconstructed context affect downstream QA performance? (\S\ref{sec:ablation})

\end{enumerate}

\subsection{Effectiveness Evaluation}
\begin{table*}[!ht]
\centering
\footnotesize
    \caption{
    \textmd{Average performance of \oreo across five QA benchmarks using Flan-T5 and OPT-IML as downstream generators, compared with baseline methods. \textit{SC} denotes Selective-Context. Performance on single-hop and multi-hop QA tasks is evaluated using Exact Match and Unigram F1, respectively. \textbf{Bold} values indicate the best results. }
    \label{tab:average}} 
\vspace{0.6em}
\begin{tabular}{lcccccccccccc}
\toprule
 & No Retrieval & Full & Passage & CXMI & SC & LLMLingua & LLMLingua-2 & xRAG & CompAct & EXIT & Refiner  & \oreo (Ours) \\ 
Task &  & & & & & & & & & & & \\
\midrule
\multicolumn{13}{c}{\textit{Flan-T5 as the downstream generator}} \\
Single-hop QA & 0.1088 & 0.3662 & 0.3394 & 0.4016 & 0.2713 & 0.3491  & 0.269 & 0.2863 & 0.3603 & 0.3487 & 0.3680 & \textbf{0.4451}\\
Multi-hop QA & 0.4485 & 0.5671 & 0.5398 & 0.603 & 0.5297 & 0.5745 & 0.5576 & 0.4828 & 0.5974 & 0.5923 & 0.5925 & \textbf{0.658} \\
\midrule
\multicolumn{13}{c}{\textit{OPT-IML as the downstream generator}} \\
Single-hop QA & 0.125 & 0.2300 & 0.2698 & 0.2714 & 0.1696 & 0.1726 & 0.2461 & 0.2297 & 0.3142 & 0.3075 & 0.3160 & \textbf{0.3616}\\
Multi-hop QA &0.4416 & 0.334 & 0.5866 & 0.4626 & 0.346 & 0.4363 & 0.5501 & 0.4818 & 0.5865 & 0.5828 & 0.5834 & \textbf{0.6542} 
\\
\bottomrule
\end{tabular}
\vspace{0.6em}
\end{table*}

\label{sec:overall}
\textbf{Overall Performance}. Table \ref{tab:average} reports the average performance across single-hop and multi-hop QA tasks using Flan-T5 and OPT-IML as downstream generators, with contexts obtained through various retrieval, extraction, and compression strategies. Across all settings, \oreo consistently outperforms other approaches, achieving the highest performance on both single-hop (Exact Match) and multi-hop (F1) tasks. Flan-T5 generally delivers superior performance compared to OPT-IML, likely due to its more advanced instruction tuning.

\begin{figure}[!htb]
%\vspace{1em}
\begin{center}
  \includegraphics[width=0.495\textwidth]{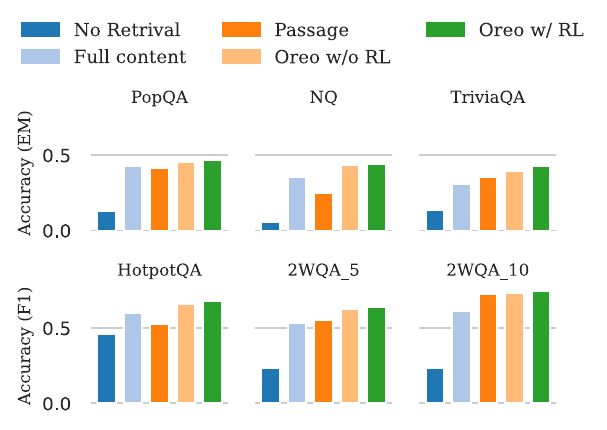}
  \end{center}
  \caption{\textmd{Performance on five datasets by using query without retrieval, original full concatenation of chunks, passage-level filtering, context generated by \oreo with and without RL. 2WQA_\textit{k} represents retrieving top-\textit{k} documents for the 2WQA dataset. The downstream generator is Flan-T5. Performance of PopQA, NQ and TriviaQA are measured by Exact Match and HotpotQA and 2WQA are measured by unigram F1. }}
  \Description{perform}
  \label{fig:metrics}
\vspace{0.6em}
\end{figure}

\textbf{Comparison against context configurations.} Figure \ref{fig:metrics} presents the performance of different setups across five datasets (with Flan-T5 as downstream generator): using query-only inputs (without retrieval), original full content, passage-level filtering, \oreo with and without RL. The results demonstrate that \oreo surpasses all other configurations across five datasets using Flan-T5. For single-hop QA tasks, \oreo achieves notable improvements in EM scores, with gains of 8.8\%, 23.1\%, and 37.5\% on the PopQA, NQ, and TriviaQA datasets compared with using original full context respectively. The relatively smaller improvement on PopQA can be attributed to the nature of its queries, which involve rare and long-tail entities. In the case of more complex multi-hop QA tasks, \oreo achieves F1 score improvements of 12.7\% on HotpotQA and 19.8\% on 2WQA. These improvements are comparatively less pronounced than those seen in single-hop tasks. This discrepancy likely stems from the increased task complexity inherent in multi-hop QA. The additional challenge of ensuring coherence in abstractive multi-hop reasoning from fragmented chunks underscores the potential for further optimization in \oreo's handling of such tasks. The experiments conducted on the 2WQA dataset using top-\textit{5} and top-\textit{10} retrieved document chunks demonstrate \oreo's flexibility in handling different input lengths. The improved performance with the top-\textit{10} chunks arises from the increased likelihood of covering more passages that contain the ground-truth answer.

Overall, these results reveals \oreo's capability to capture essential information and filter out distracting content from retrieved document chunks, leading to improved performance in downstream factual QA tasks. The modest improvements achieved with RL further emphasize its value in addressing knowledge inconsistencies between the retriever and generator. 

\begin{figure*}[!htb]
\begin{center}
  \includegraphics[width=0.936\textwidth]{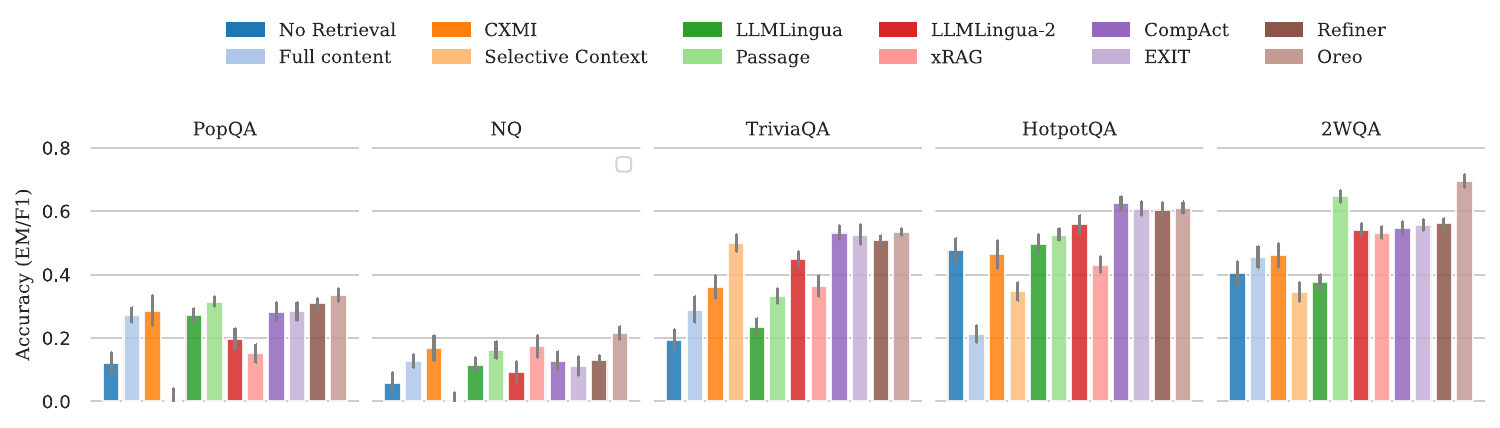}
  \end{center}
  \caption{\textmd{Performance comparison with 95\% confidence intervals against baselines using OPT-IML as the generator. Specifically, \textit{Passage} denotes passage-level filtering, \textit{CXMI} refers to filtering guided by conditional cross-mutual information, and \textbf{Full} represents the use of original content without any filtering. PopQA, NQ, and TriviaQA are evaluated with Exact Match scores, while HotpotQA and 2WQA use Unigram F1 for accuracy measurement}}
  \Description{perform}
  \label{fig:methods}
\end{figure*}

\textbf{Comparison against baselines. }
We compare the quality of generated context by \oreo against a suite of representative extraction and compression methods across five QA datasets. Table~\ref{tab:main} summarizes the performance and token counts across five datasets by employing Flan-T5 as the downstream generator. From the table, it is evident that \oreo outperforms almost all selected methods across five datasets, with improvements ranging from 0.35\% to 8.58\% over the second-best methods. These gains are particularly pronounced in extractive, single-hop tasks such as NQ and PopQA, where concise yet precise evidence retrieval is paramount. On multi-hop tasks, \oreo remains competitive but shows relatively smaller gains, as these tasks require complex evidence chaining and reasoning to synthesize evidence scattered across multiple chunks. In addition to achieving superior performance, \oreo significantly reduces the context length provided to the downstream generator while maintaining or even enhancing task accuracy. We further validate these findings by evaluating with OPT-IML as the downstream generator (see Fig.~\ref{fig:methods}). Consistent with results of using Flan-T5, \oreo leads to the highest performance across four datasets (except HotpotQA), bringing +0.0211 EM (PopQA), +0.0405 EM (NQ), +0.0019 EM (TriviaQA), -0.0142 F1 (HotpotQA) and +0.0495 F1 (2WQA) improvements compared with the second-best baseline method. 

\textbf{SOTA methods observations.} Among selected SOTA methods, CompAct \cite{yoon-etal-2024-compact} marginally outperforms \oreo on HotpotQA, achieving a 2.26\% higher F1 score. This advantage is attributed to CompAct’s incremental and iterative compression strategy, which proves beneficial for tasks requiring deep multi-hop reasoning. However, its inference latency is nearly 3X that of \oreo, presenting a trade-off between accuracy and efficiency. Other strong performers include CXMI-guided model, Refiner and EXIT, which use query-aware or contrastive objectives to maintain relevance. In contrast, Selective-Context yields the weakest results overall. This underperformance likely stems from its reliance on self-information of lexical units (\textit{eg.}, tokens, phrases, or sentences), which fail to capture dependencies among semantic units. Similarly, LLMLingua and LLMLingua-2, which employ perplexity-based filtering, also struggle across datasets. Their reliance on self-information and perplexity metrics, without explicit query conditioning, limits their ability to extract context tightly aligned with user queries. 

\begin{table*}
    \caption{\textmd{Summary of QA task performance and token counts using context derived from different methods. Flan-T5 is the generator. Performance on PopQA, NaturalQuestions, and TriviaQA is evaluated using Exact Match, while HotpotQA and 2WikiMultihopQA are assessed using Unigram F1. \textbf{Bold} values indicate the best performance among all methods, \textit{italics} text denotes the second-best performance. The values in parentheses indicate the percentage improvement of the best-performing method over the second-best method. All datasets are tested with top-\textit{5} retrieved chunks and all retrieved passages are set the same for different methods. }}
  \centering
   \resizebox{\linewidth}{!}
   {
    \begin{tabular}{lcccccccccc}
    \toprule
    \multirow{2}[3]{*}{\textbf{Methods}} & \multicolumn{2}{c}{\textbf{PopQA}} & \multicolumn{2}{c}{\textbf{NaturalQuestions}} & \multicolumn{2}{c}{\textbf{TriviaQA}} & \multicolumn{2}{c}{\textbf{HotPotQA}} & \multicolumn{2}{c}{\textbf{2WikiMultihopQA}} \\
\cmidrule(lr){2-3}\cmidrule(lr){4-5}\cmidrule(lr){6-7}\cmidrule(lr){8-9}\cmidrule(lr){10-11}         & \textbf{EM} & \textbf{\# tokens} & \textbf{EM} & \textbf{\# tokens} & \textbf{EM} & \textbf{\# tokens} & \textbf{F1} & \textbf{\# tokens} & \textbf{F1} & \textbf{\# tokens} \\
    \midrule
    No Retrieval      & 0.1320         & 30           & 0.0558                    & 39           & 0.1387                    & 31           & 0.4599                    & 47           & 0.4371                     & 35           \\
Full content      & 0.4305                     & 1689         & 0.3584                    & 1636         & 0.3097                    & 1676         & 0.6014                    & 1707         & 0.5328                     & 1786         \\
CXMI              & 0.4202                     & 340          & 0.3917                    & 329          & 0.3929                    & 354          & 0.6409                    & 351          & 0.565                      & 305          \\
Selective Context & 0.1445                     & 199          & 0.3981                    & 193          & 0.2712                    & 203          & 0.5588                    & 214          & \textit{0.6106}            & 158          \\
LLMLingua         & 0.2702                     & 497          & \textit{0.4125}           & 491          & 0.3647                    & 520          & 0.5584                    & 527          & 0.5905                     & 394          \\
Passage           & 0.4150                     & 131          & 0.2506                    & 183          & 0.3526                    & 203          & 0.5280                    & 190          & 0.5515                     & 205          \\
LLMLingua-2       & 0.2603                     & 252          & 0.1892                    & 247          & 0.3573                    & 265          & 0.6186                    & 269          & 0.4965                     & 279          \\
xRAG              & 0.2117                     & 31           & 0.2580                    & 40           & 0.2893                    & 32           & 0.4777                    & 48           & 0.4879                     & 36           \\
CompAct           & 0.3917                     & 142          & 0.265                     & 173          & \textit{0.4242}           & 180          & \textbf{0.6932}           & 174          & 0.5015                     & 173          \\
EXIT              & 0.3972                     & 124          & 0.2301                    & 211          & 0.4188                    & 208          & 0.6742                    & 180          & 0.5104                     & 123          \\
Refiner           & \textit{0.4312}            & 91           & 0.2677                    & 148          & 0.4051                    & 148          & 0.6706                    & 131          & 0.5144                     & 102          \\
\textbf{\oreo (Ours) }               & \textbf{0.4682 (+ 8.58\%)} & \textbf{108} & \textbf{0.4413 (+6.98\%)} & \textbf{134} & \textbf{0.4257 (+0.35\%)} & \textbf{130} & \textit{0.6775 (-2.26\%)} & \textbf{272} & \textbf{0.6384 (+ 4.55\%)} & \textbf{103}\\
    \bottomrule
    \end{tabular}
    }
  \label{tab:main}
\end{table*}

\subsection{Efficiency Evaluation}
\label{sec:latency}
We assess the efficiency of \oreo from two key perspectives: (1) the trade-off between context length reduction and downstream QA performance, and (2) the trade-off between end-to-end inference latency and QA performance. 

\begin{figure}[H]
\vspace{-1.em}
\begin{center}
  \includegraphics[scale=0.7]{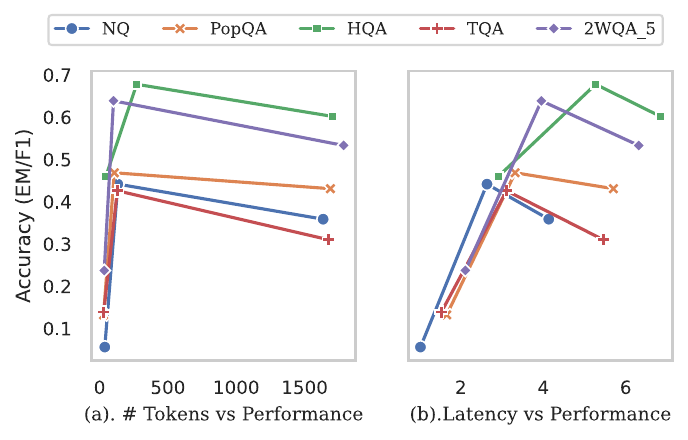}
  \end{center}
 % \vspace{-1em}
  \caption{\textmd{ Left (a) - Comparison of number of input tokens for generator and QA performance across different context types. Right (b) - Comparison of end-to-end inference time (measured in seconds) by using different types of context. }}
  \Description{token_vs_perf}
  \label{fig:token_perf}
  \vspace{-1.em}
\end{figure}

Figure~\ref{fig:token_perf} illustrates the number of tokens forwarded to the downstream generator and the total inference latency, which includes both \oreo’s context reconstruction and the subsequent generation time. We compare three input configurations: query-only (no retrieval), full document content, and the context reconstructed by \oreo. \oreo achieves a substantial reduction in input length, compressing the context by 84\% to 94\% compared to full document content. This compression is accompanied by a latency reduction of 22.98\% to 43.01\%, while simultaneously delivering significant performance improvements ranging from 8.76\% to 37.46\%. These gains are especially pronounced in extractive QA tasks (\textit{eg.}, NQ, TriviaQA) as shown from the steeper improvement in Figure \ref{fig:token_perf} from right endpoints to peaks. The high compression rate and improved performance demonstrates \oreo’s capability to effectively condense the retrieved context by preserving only the most critical evidence required for accurate answer generation. This also indicates the context reconstructed by \oreo is highly utilized by the downstream generator. The favorable trade-off between latency and performance underscores \oreo’s potential for real-world applications, offering both computational efficiency and improved task accuracy for scalable, high-throughput RAG systems.

\subsection{Robustness Evaluation}
We evaluate \oreo's robustness from two aspects: its sensitivity to irrelevant or distracting information (noise robustness), and its ability to handle arbitrary rankings of retrieved chunks (order robustness).

\begin{figure}[hb]
\begin{center}
  \includegraphics[scale=0.8]{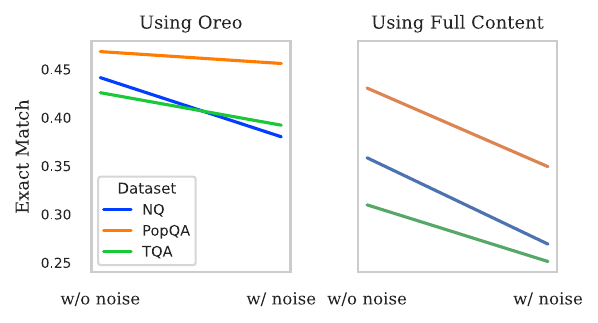}
  \end{center}
  \caption{\textmd{Performance declines as irrelevant chunks are introduced into the retrieved chunk set. }}
  \Description{npose}
  \label{fig:noise}
\end{figure}

\textbf{Noise robustness. }
We evaluate the robustness of \oreo in handling noise within the retrieved documents, focusing specifically on extractive QA tasks. In this evaluation, we retain a single chunk that explicitly contains the ground-truth answer and introduce four irrelevant documents to simulate a noisy retrieval scenario. This setup examines \oreo's effectiveness in filtering out distractions content and identifying query-specific information to generate accurate responses. Figure \ref{fig:noise} depicts the performance degradation as irrelevant chunks are added to the context. Compared to directly concatenating all retrieved chunks as context, context reconstructed by \oreo demonstrates a smaller decrease in EM scores and a slower rate of decline, as evidenced by a less steep slope.

%\label{sec:effect}

\textbf{Order robustness. } 
\label{sec:robust}
We evaluate the robustness of \oreo to variations in the order of retrieved documents by shuffling the top-\textit{5} retrieved results and comparing its performance against the original document order. The results for five datasets are presented in Table \ref{tab:shuffle}. From the table we can see that, \oreo consistently maintain the performance on five datasets (with $\pm 0.003$ to $\pm 0.027$). This highlights that \oreo is order- or position-agnostic. Even the retrieved chunks are suboptimally ranked or presented in an arbitrary order, \oreo can still effectively capture and synthesize essential information as long as the evidence exists in the chunks. This capability is largely attributed to \oreo's inherent reordering feature during context reconstruction, enabling it to function as an implicit reranker. Such robustness is particularly valuable for mitigating the "lost-in-the-middle" \cite{liu2023lost} phenomenon, where the order of relevant information may influence the downstream generator's performance. 

\subsection{Generalizability Evaluation}
%in-domain vs. out-of-domain
\label{sec:zeroshot}

To evaluate the cross-dataset generalizability of \oreo, we assess its transferring capability by applying models trained on one dataset to tasks in a different dataset without any fine-tuning. This approach tests \oreo's ability to generalize its context reconstruction and synthesis capabilities to unseen query distributions. Specifically, we examine performance when using a model trained on PopQA to generate answers for NQ and a model trained on 2WQA to process HotpotQA queries. We report the detailed results in Table \ref{tab:zero-shot} in Appendix~\ref{app:general}, which demonstrate that \oreo achieves competitive performance in the zero-shot setting. For example, the model trained on PopQA achieves a score of 0.4352 when applied to NQ, only slightly lower than the performance of being specific trained (\textit{ie.} 0.4413 and 0.4682). Similarly, using the 2WQA-trained model on HotpotQA yields a score of 0.6344, closely matching the intra-dataset score of 0.6384. These findings demonstrate \oreo's ability to generalize its context reconstruction to similar QA types effectively, even under query distribution shifts. Its strong performance across datasets highlights its robustness and adaptability, making it a promising solution for open-domain QA tasks that require flexibility in handling diverse knowledge sources and query structures.

\begin{table}[H]
\caption{\textmd{QA performance when shuffling the retrieved documents. }}
\vspace{0.5em}
  \centering
    \begin{tabular}{lcc}
    \toprule
     \textbf{Dataset} & \textbf{w/o shuffle} & \textbf{w/ shuffle}\\
    \midrule
    PopQA  & 0.468 & 0.441 \\
    NaturalQuestions  & 0.441 & 0.425 \\
    HotpotQA  & 0.426 & 0.429 \\
    TriviaQA  & 0.678 & 0.668 \\
    2WikiMultihopQA  & 0.638 & 0.614 \\
    \bottomrule
    \end{tabular}
  \label{tab:shuffle}%
  \vspace{-0.5em}
\end{table}%

\subsection{Faithfulness and Completeness Evaluation}
\label{sec:qwen_eval}
Apart from the downstream task performance, the quality of context generated by \oreo is essential, esp. the factual accuracy (faithfulness) and coverage of relevant information (completeness) with respect to the original retrieved passages. To this end, we conduct an evaluation of both faithfulness and completeness to ensure that \oreo produces context that is not only concise but also reliable and fully representative of the source passages. 

We adopt the LLM-as-a-judge framework \cite{gu2024survey} to systematically assess these dimensions. In particular, we prompt Qwen2.5-Instruct \cite{yang2024qwen2} to evaluate the generated context by assigning faithfulness and completeness scores on a 0–10 scale. \textit{Faithfulness} reflects the degree to which the context remains factually grounded in the original retrieved content, avoiding hallucinations or the introduction of extraneous information. \textit{Completeness} assesses whether the context sufficiently captures all salient and relevant details from the original passages with respect to the query. To promote transparency and interpretability, the model is also asked to provide a rationale supporting each score. Prompts for scores and explanation generation are provided in Appendix~\ref{app:llmjudge}.

We provide evaluation results in Figure~\ref{fig:qwen_eval} in Appendix~\ref{app:qwen_results}
Among all methods, \oreo consistently achieves the highest scores across all datasets, excelling in both completeness and faithfulness, with standout scores on complex datasets like HotpotQA (8.87 completeness, 8.97 faithfulness). CompAct emerges as a strong second-best, showing strong balance and high faithfulness, especially on HotpotQA and 2WQA. Refiner delivers moderately competitive results, generally maintaining factual consistency but showing limitations in coverage. EXIT demonstrates lower overall performance, especially struggling on more demanding datasets such as 2WQA. In contrast, LLMLingua and LLMLingua-2 produce the weakest results, with both completeness and faithfulness significantly lower across all datasets, likely due to aggressive filtering or compression strategies that sacrifice critical information.

\subsection{Ablation Study}
\label{sec:ablation}
We conduct an ablation study to investigate the effect of varying context lengths generated by \oreo. Specifically, we progressively increase the minimum token threshold for context generation from 30 to 300 tokens, in increments of 30, while fixing the number of retrieved passages to top-\textit{5}. The results of downstream task performance are summarized in Figure~\ref{fig:ablation} in Appendix~\ref{app:ablation}. Our findings indicate a consistent performance improvement across all datasets as the minimum context length increases, with gains being more pronounced in the earlier stages (from 30 to 180 tokens). These improvements suggest that extending the context allows the model to access a broader set of relevant evidence, improving its ability to synthesize accurate responses. However, beyond a threshold, typically between 240 and 270 tokens, we observe a performance plateau or marginal decline. This indicates diminishing returns with excessively long contexts. While longer windows can potentially capture more relevant details, they also risk introducing extraneous, redundant, or weakly relevant content, which can dilute the core information necessary for accurate answers.

\section{Related Work}
\subsection{Post-retrieval Enhancement for RAG}
Post-processing methods are widely employed to refine retrieved content for improved downstream generation. These methods can be categorized as follows:

\textbf{Reranking}. Rerankers reorder and prioritize retrieved documents to emphasize the most relevant results. They typically operate sequentially or iteratively after retrieval, leveraging various criteria such as semantic relevance between query and passages \cite{glass-etal-2022-re2g, hofstatter2023fid}, connections among documents \cite{dong2024don}, the majority of reader predictions \cite{mao-etal-2021-reader}, and utility for generation \cite{ma-etal-2023-large}. Rerankers are usually based on cross-encoder (\textit{eg.}, BGE \cite{bgeembedding}, Mixedbread \cite{li2023angle}), multi-vector models (\textit{eg.}, ColBert \cite{khattab2020colbert, santhanam-etal-2022-colbertv2}). Recent works also explore using LMs as rerankers (\textit{eg.}, RankT5 \cite{zhuang2023rankt5}, RankZephyr \cite{pradeep2023rankzephyr}, RankGPT \cite{sun-etal-2023-chatgpt}, DPA-RAG \cite{dong2024understand}). %DPA-RAG \cite{dong2024understand} integrates pair-wise, point-wise, and contrastive preference alignment into the reranker, enabling external preference alignment across the components of the RAG system.

% \textbf{Filtering}. Filtering is the process of selectively removing irrelevant, redundant, or low-quality information from retrieved documents to enhance their usefulness. Filtering can be performed at different granularity levels such as passage-level \cite{asai-etal-2022-evidentiality, yoran2023making, asai2023self}, sentence-level \cite{wang2023learning, hwang-etal-2024-dslr} or finer token-level \cite{liu-etal-2023-tcra, jin-etal-2024-bider, anderson-etal-2022-lingua, jiang-etal-2024-longllmlinguaCompAct, mao2024fit}. self-RAG filters out irrelevant passages by self-reflection mechanism. FIT-RAG \cite{mao2024fit} retrieves and compresses top-\textit{10} documents by extracting the sub-documents.

 \textbf{Post verification and correction}. Some studies incorporate post-hoc evaluations to improve factuality and relevance of retrieved documents. Examples include relevance evaluators \cite{yan2024corrective}, fact-checkers \cite{liu2306reta}, attribution \cite{gao-etal-2023-rarr, yu-etal-2024-chain} and multi-agent~\cite{xu2024activerag} mechanisms to further solidify the accuracy and reliability of the retrieved documents and responses. 
 
 % For example, \cite{yan2024corrective} uses a retrieval evaluator accessing the relevance of the retrieved documents. RETA-LLM \cite{liu2306reta} includes a fact checking module to verify whether there exist factual errors in the generated answers. CHAIN-OF-NOTE (CON) \cite{yu-etal-2024-chain} accesses relevance of each retrieved document to the query by generating sequential reading notes. ACTIVERAG \cite{xu2024activerag} involves multiple agents to build coherent knowledge understanding and refine the internal reasoning of LLMs. While it achieves strong performance, the multi-agent communication process requires three separate LLM inference steps to generate a final response, leading to higher latency and increased API costs.
 
 \textbf{Compressing}. Compression methods condense retrieved content to improve efficiency and focus. These methods can be broadly categorized into lexical-based and embedding-based approaches. Lexical-based methods usually involve summarization techniques \cite{xu2024recomp, liu-etal-2023-tcra} to retain essential information, semantic filtering to remove low-importance tokens, and both extractive and abstractive strategies for eliminating irrelevant context \cite{xu2024recomp}. Some approaches compute the self-information of lexical units to discard less informative content \cite{li-etal-2023-compressing}, or apply token-level filtering based on perplexity \cite{jiang-etal-2024-longllmlinguaCompAct}. Embedding-based methods, on the other hand, condense documents into fixed-size representations in the embedding space, recent works include xRAG \cite{cheng2024xrag} and PISCO \cite{louis2025pisco}.
 Our work falls falls within the lexical-based compression group. 
 
 %A variety of techniques have been proposed in this space. TCRA-LLM \cite{liu-etal-2023-tcra} employs a semantic compressor to filter out low-importance tokens for food recommendation tasks. RECOMP \cite{xu2024recomp} introduces dual-encoder compressors for both extractive and abstractive compression to eliminate irrelevant context. Selective-Context \cite{li-etal-2023-compressing} uses a lightweight LM to compute the self-information of lexical units and discard less informative content. LongLLMLingua \cite{jiang-etal-2024-longllmlinguaCompAct} performs coarse-grained compression using document-level perplexity and fine-grained token-level filtering based on token perplexity. Unlike the above works, embedding-based methods such as xRAG \cite{cheng2024xrag} and PISCO \cite{louis2025pisco} employs soft prompt compression to reduce the length of multiple documents. 

\subsection{Reinforcement Learning for Large Language Models}
Reinforcement learning for Language Models (RL4LM) has emerged as a transformative technique to further enhance LLMs' performance during the post-training process \cite{cao2024survey, pternea2024rl}. Traditional RL4LM usually involves a reward model, for example using PPO \cite{schulman2015high} to update the policy model (\textit{eg.}, InstructGPT \cite{ouyang2022training}, GPT-4 \cite{achiam2023gpt}). Some RL4LM such as Direct Preference Optimization (DPO) \cite{rafailov2024direct}
and Reward-aware Preference Optimization (RPO) \cite{adler2024nemotron} get rid of the reward model to provide more stable and computationally efficient solutions (\textit{eg.}, Qwen 2 \cite{chu2024qwen2} and  Nemotron-4 \cite{adler2024nemotron}). Common goals of RL4LM include improving performance of downstream NLP tasks \cite{deng-etal-2022-rlprompt, ghalandari-etal-2022-efficient, ramamurthy2022reinforcement}, minimizing data and resource dependencies \cite{zhang2022tempera}, aligning model outputs with user intent, values and goals \cite{ouyang2022training}, and adhering to responsible AI principles \cite{bai2022training, bai2022constitutional}. Human feedback can be integrated into the framework by constructing preference datasets, which are then used to fine-tune both the policy and reward models (also termed as Reinforcement Learning from Human Feedback (RLHF)) \cite{bai2022training, ouyang2022training, hu2023aligning}. Some studies also explore RL4LM without human feedback \cite{rafailov2024direct} or replaced with AI feedback \cite{bai2022constitutional, yuan2024self} by distillation from LLMs \cite{cui2023ultrafeedback, park-etal-2024-offsetbias}, prompting LLMs as reward functions \cite{kwon2023reward, lee2023rlaif, zhang2024generative}, and self-rewarding \cite{yuan2024self}, or using performance-based metrics such as fluency or coherence \cite{ghalandari-etal-2022-efficient}, and task-specific constraints over the distribution of language \cite{ramamurthy2022reinforcement, wu2024beta}. In the specific domain of RAG, RRAML \cite{bacciu2023rraml} employs RL to train a retriever in arbitrarily large databases. PRCA \cite{yang-etal-2023-prca} applies RL to fine-tune the context to optimize the reward for the generator. BIDER \cite{jin-etal-2024-bider} adopts RL to bridge the inconsistency between the retriever and generator. 

\section{Conclusion}
We have presented \oreo - a lightweight and pluggable module designed to enhance the performance of RAG systems by reconstructing retrieved document chunks and mitigating the potential knowledge inconsistencies between the retriever and generator. Upon receiving document chunks from the retriever, \oreo efficiently filters out irrelevant, redundant and distracting content, transforming them into a concise and query-supportive context. These reconstructed contexts effectively guide the generator toward producing accurate answers for open-domain QA tasks. Notably, \oreo can be seamlessly integrated with arbitrary retrievers, generators, or other RAG components without requiring significant adjustments or modifications. Experimental results demonstrate \oreo's effectiveness in downstream tasks, its efficiency in compressing context while improving performance, and its robustness in handling noisy and imperfectly ranked document chunks. 

\textbf{Limitations.}
While \oreo shows strong performance on open-domain QA tasks, it has some limitations. First, its aggressive compression may omit essential information in complex settings like multi-hop or long-form QA. Second, \oreo has not been systematically tested in adversarial retrieval scenarios~\cite{wang2025retrieval} involving conflicting or deceptive content. Third, current evaluations rely heavily on indirect metrics such as downstream QA accuracy or LLM judgments, which may introduce bias. 

\textbf{Future Work.} 
Future research will explore adaptive compression strategies that dynamically allocate token budgets based on query complexity and task type. Robustness to adversarial or noisy retrieval scenarios also warrants closer investigation, especially for high-stakes domains. 
We are also interested in developing more principled and fine-grained evaluation frameworks to better understand the trade-offs between compression, informativeness and faithfulness in context reconstruction. Lastly, to address
sparsity in rewards, a promising direction for future work is to develop progress-based RL frameworks that incorporate intermediate quality assessments of \oreo's reconstructed context, providing denser and more fine-grained rewards to enable more stable and efficient policy learning.

%\section{Acknowledgments}
%\iffalse 
\appendix
\section{Prompt Templates for Data Collection}
\label{app:prompts}
\subsection{Prompt Templates for Data Collection}
\label{tab:data_gen}
%\begin{table}[htb]
%\caption{Prompt template used for data collection.\label{tab:data_gen}}
%\begin{prompt}[title={Prompt}, label=prompt:data_gen]
\fbox{\begin{minipage}{26em}
{\bf Input:}
Your task is to decompose the question, extract and abstract supporting information from the context to answer the question. Your output should mention all entities involved in the question, supporting sentences and rationals to all sub-questions from the context. If the conetxt doesn't provide information to answer the question, output '[UNKNOWN]'. Output the <Output> part only.\\
Example1:\\
    <Question>: Where was the director of film The Fascist born? \\
    <Context>: \{Retrieved document chunks\} \\
    <Output>: Luciano Salce, the director of the satirical film The Fascist, was born on September 25, 1922, in Rome, Italy. Salce was an Italian filmmaker, actor, and screenwriter known for his ability to blend comedy with social and political critique. \\
    Example2: \\
    <Question>: what is the number 1 sport in the usa? \\
    <Context>: \{Retrieved document chunks\} \\
    <Output>: American football is the most popular sport in the United States followed by basketball, baseball, and soccer. \\
    Example3:  \\
    <Question>: What was the first English monastery to be sacked by the Norsemen?\\
    <Context>: \{Retrieved document chunks\} \\
    <Output>: Vikings attacked the monastery at Lindisfarne on June 8, 793, which is the first recorded Viking raid on an English monastery.\\
    Example4:\\
    <Question>: Kate Philips played which wife of Henry VIII in 'Wolf Hall'? \\
    <Context>: \{Retrieved document chunks\} \\
    <Output>: Kate Phillips played Abigail Williams in "The Crucible" at the West Yorkshire Playhouse, and then went on to film her scenes for the BBC's adaptation of "Wolf Hall" in which she played Jane Seymour, Henry VIII's third wife. \\
    Example5:\\
    <Question>: Lokomotiv Yaroslavl was the team founded in 2011 after the plane crash near which airport? \\
    <Context>: \{Retrieved document chunks\} \\
    <Output>: Lokomotiv Yaroslavl Hockey Club Lokomotiv, also known as Lokomotiv Yaroslavl, is a Russian professional ice hockey team. On 7 September 2011, nearly the entire team perished in the Lokomotiv Yaroslavl plane crash. The aircraft ran off the runway before lifting off, struck a tower mast, caught fire and crashed from the end of the runway of Tunoshna Airport on the Volga River bank. \\

\{Question\} \\ 
\{Retrieved document chunks\} \\
{\bf Output:} \{Output\}
%\end{prompt}
\end{minipage}}
%\end{table} 

\subsection{Prompt Template for Boostraping Data Generation}
\label{tab:boost_gen}
%\begin{table}[H]
%\caption{Bootstraping Prompt Template \label{tab:boost_gen}}
%\begin{prompt}[title={Prompt}, label=prompt:boost_gen]
\fbox{\begin{minipage}{26em}
{\bf Input:}
You are given a question, a set of document chunks, a correct answer, extract evidences and supporting information from the chunks and generate rationales how these information derive the correct answer.
\\
Example1:\\
    <Question>: What nationality were social anthropologists Alfred Gell and Edmund Leach? \\
    <Chunks>: \{Retrieved document chunks\} \\
    <Correct answer>: British. \\
    <Output>: Both Alfred Gell and Edmund Leach were British. They were educated and primarily worked within the United Kingdom's academic framework. Their national and professional affiliations firmly establish their British nationality. \\
    Example2: \\
    <Question>: Crucible is a geodemography computer system created by a company that has stores in how many countries? \\
    <Chunks>: \{Retrieved document chunks\} \\
    <Correct answer>: 12. \\
    <Output>: Crucible is a geodemography computer system created by Tesco, a multinational grocery and general merchandise retailer. Tesco has stores in 12 countries as of recent data, so 12 is the answer. \\
    Example3: \\
    <Question>: What word is in both the genre of Muhammed Suiçmez's band and the genre of Dave Meniketti's band? \\
    <Chunks>: \{Retrieved document chunks\} \\
    <Correct answer>: Metal. \\
    <Output>: Necrophagist is known for its death metal style. Y\&T is often classified under the broader category of heavy metal. So the answer is mental. \\
    
\{Question\} \\ 
\{Retrieved document chunks\} \\
\{Correct answer\} \\
{\bf Output:} \{output\}
%\end{prompt}
\end{minipage}}
%\end{table} 
\section{Statistics and Experimental Setups for Datasets}
\label{app:stats_dataset}
Table \ref{table:stat} provides detailed statistics for each dataset, including the number of samples in the training set after curation, test set size, the specific retriever used, and evaluation metrics. Besides, we use the precision\textit{@k} as an approximation of retrievers' performance. Precision\textit{@k} is defined as the ratio of chunks that contain the among all retrieved chunks \textit{k} for each query.  

\begin{table*}[!htb]
\caption{\textmd{Dataset statistics, retrievers and evaluation metrics. EM -Exact Match, F1 - Unigram F1}}
\label{table:stat}
\begin{tabular}{ c|cccccc } 
 \hline
 \textbf{Dataset} & \textbf{\# Train (k)} & \textbf{\# Test (k)} & \textbf{Retriever} & \textbf{Precision@\textit{5}} & \textbf{Task} & \textbf{Metric} \\ 
  \hline 
PopQA & 6.5 & 1.4 & Contriver & 0.287 & Extractive single-hop QA & EM \\ 
 NaturalQuestions & 28.3 & 3.6 & DPR & 0.33 & Extractive single-hop QA & EM\\ 
 TriviaQA & 30.1 & 11.3 & Contriver & 0.43 & Extractive single-hop QA & EM\\ 
HotpotQA & 20.7 & 5.6 & Contriver & 0.137 & Abstractive multi-hop QA & F1 \\
2WikiMultiHopQA & 20.7 & 12.6 & BM25 & 0.07 & Abstractive multi-hop QA & F1\\
 \hline
\end{tabular}
\end{table*}

\section{Parameter Settings}
\label{app:params}
We detail the key hyperparameters and configurations used across all experiments in Table~\ref{table:params}. Specifically, \textit{CML} and \textbf{RL} represents contrastive multitask learning and reinforcement learning respectively. 

\begin{table}[H]
\centering
\caption{\textmd{Parameter settings for experiments. Parameters without being specified are set to their default values as defined by the development package. }}
\vspace{0.6em}
\begin{tabular}{c|c} 
 \hline
 Parameter & Value \\  
 \hline
    $\eta$ (CML) & 0.01\\
    $\alpha$ (CML) & 0.5\\
    $\epsilon$ (RL) & 0.2 \\
    $\gamma$, $\lambda$(RL) & 0.95 \\
    Top-\textit{k} (RL) & 4 \\
    Top-\textit{p} (RL)& 0.95 \\
 \hline
\end{tabular}
\label{table:params}
\end{table}

\section{LLM-as-a-Judge For Faithfulness and Completeness Evaluation}
\label{app:llmjudge}
\subsection{Prompts for Qwen-2.5-Instruct}
To directly evaluate the quality of context generated by \oreo, we employ Qwen-2.5-Instruct \cite{yang2024qwen2} as a reference model to assess two critical dimensions: faithfulness - how well the answer aligns with the retrieved passages, and completeness - to what extend does the generated context cover all essential information to correctly answer the query.

We design instructions for prompting Qwen-2.5-Instruct, as detailed in Table~\ref{tab:qwen_judge_faith} and Table~\ref{tab:qwen_judge_comp}.
\begin{table}[h]
\caption{Qwen-2.5-Instruct used for faithfulness evaluation \label{tab:qwen_judge_faith}}
\begin{prompt}[title={Faithfulness evaluation prompt}, label=judge-llm]
{\bf Input:}
"""You are an expert evaluator assessing the **faithfulness** of a generated context with respect to the original passages. You will be given a query, a set of original passages, and a generated context.

Your task is to determine how accurately the generated context reflects the facts and meanings in the original passages, focusing only on the information required to answering the query. Consider whether the context includes hallucinated information, omits key facts, or misrepresents any content.

Rate the faithfulness on a scale from 0 to 10:
- 0: The generated context is entirely unfaithful or unrelated to the original passages.
- 10: The generated context is fully faithful, with no hallucinations, distortions, or omissions of relevant information.

Output your score (a float between 0 and 10), followed by a concise explanation of your reasoning. """\\
{\bf Text:} \{text\}\\
{\bf Output:} \{Output\}
\end{prompt}
\end{table} 

\begin{table}[h]
\caption{Qwen-2.5-Instruct used for completeness evaluation \label{tab:qwen_judge_comp}}
\begin{prompt}[title={Completeness evaluation prompt}, label=complete]
{\bf Input:}
"""You are an expert evaluator assessing the **completeness** of a generated response. You will be given a query, a set of original passages, and a generated context intended to answer the query.

Your task is to rate how thoroughly the generated context covers all necessary information from the original passages required to answer the query. The context should not omit relevant details, and should avoid adding any external or fabricated content.

Rate the completeness on a scale from 0 to 10:
- 0: The generated context provides no useful information for answering the query.
- 10: The generated context includes all necessary information to fully and correctly answer the query.

Output your score (a float between 0 and 10), followed by a brief explanation of your reasoning. """  
\\
{\bf Text:} \{text\}\\
{\bf Output:} \{Output\}
\end{prompt}
\end{table}

\subsection{Scoring Results}
\label{app:qwen_results}
\begin{figure*}[!htb]
\begin{center}
  \includegraphics[width=0.98\textwidth]{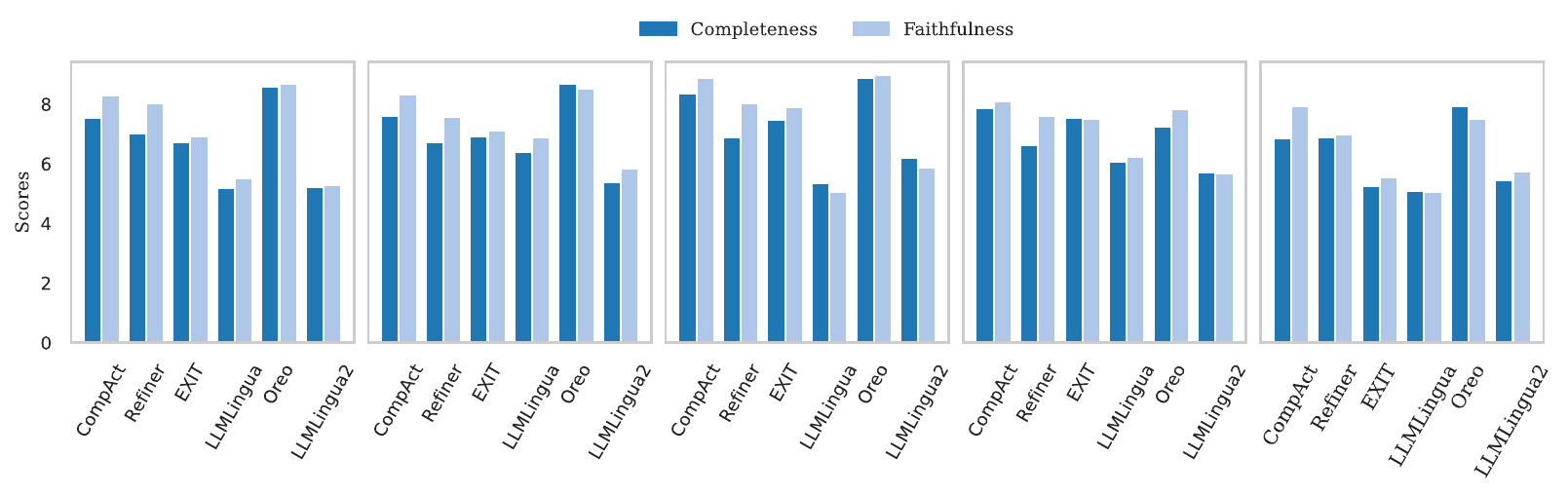}
  \end{center}
  \caption{\textmd{Completeness and faithfulness evaluation by Qwen-2.5-Instruct }}
  \Description{qwen_eval}
  \label{fig:qwen_eval}
\end{figure*}

Figure~\ref{fig:qwen_eval} presents the completeness and faithfulness scores evaluated by Qwen-2.5-Instruct, demonstrating that \oreo achieves the highest performance on both metrics.

\section{Generalizability Evaluation}
\label{app:general}
To evaluate cross-dataset generalizability, we test \oreo's transferability by applying models trained on one dataset to a different one without fine-tuning. This assesses \oreo’s ability to reconstruct and synthesize context under unseen query distributions. Specifically, we evaluate models trained on PopQA for NQ, and on 2WQA for HotpotQA. Detailed results are provided in Table~\ref{tab:zero-shot}.
\begin{table}[H]
\caption{\textmd{QA performance with zero-shot setting. PopQA $\to$ NQ represents the model trained on PopQA is applied to NQ.}}
  \centering
  % \resizebox{\linewidth}{!}{
    \begin{tabular}{ccc}
          \toprule
          \textbf{Dataset} & \textbf{Model $\to$ Dataset}  & \textbf{Performance} \\ \hline
     \multirow{3}[3]{*}{NQ} & NQ $\to$ NQ & 0.4413 \\
     & PopQA $\to$ PopQA & 0.4682 \\
     & PopQA $\to$ NQ & 0.4352 \\
    \midrule
    \multirow{3}[3]{*}{HotpotQA} & HotpotQA $\to$ HotpotQA & 0.6775 \\
     & 2WQA $\to$ 2WQA & 0.6384 \\
     & 2WQA $\to$ HotpotQA & 0.6344 \\
    \bottomrule
    \end{tabular}
  \label{tab:zero-shot}%
\end{table}%
\section{Ablation Study on Generated Token Numbers}
\label{app:ablation}
\begin{figure}[!htb]
\begin{center}
  \includegraphics[scale=0.7]{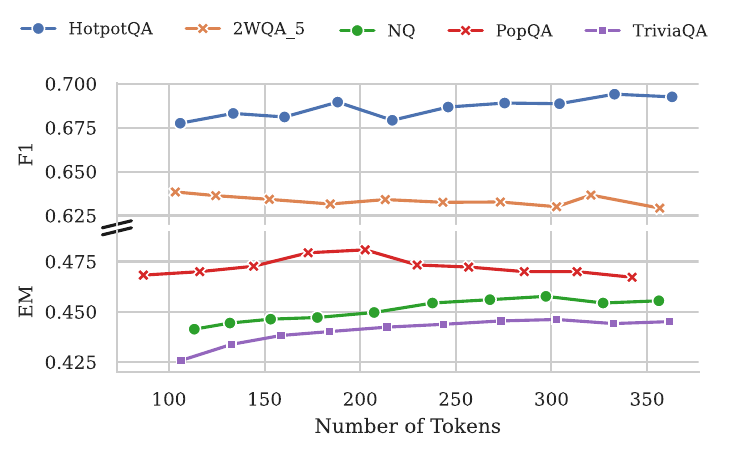}
  \end{center}
  \caption{\textmd{Performance of \oreo generating different lengths of context across five datasets }}
  \Description{ablation}
  \label{fig:ablation}
\end{figure}

We perform an ablation study to assess how varying \oreo's context length affects downstream performance. By increasing the minimum token threshold from 30 to 300 (in steps of 30) while keeping the top-5 retrieved passages fixed, we observe performance trends summarized in Figure~\ref{fig:ablation}.

%%
%% The acknowledgments section is defined using the "acks" environment
%% (and NOT an unnumbered section). This ensures the proper
%% identification of the section in the article metadata, and the
%% consistent spelling of the heading.

\newpage
%%
%% The next two lines define the bibliography style to be used, and
%% the bibliography file.
\bibliographystyle{ACM-Reference-Format}
\bibliography{paper}
\end{document}